%% file: main.tex
\documentclass[11pt]{article}
\usepackage[margin=1in]{geometry}
\usepackage{parskip}
\usepackage[utf8]{inputenc}
\usepackage{amsmath,amssymb,amsfonts,amsthm,mathtools}
\usepackage{bm,bbm}
\usepackage{graphicx,booktabs,subcaption,xcolor}
\usepackage{algorithm,algpseudocode}
\usepackage{tikz,array}
\usepackage{etoolbox}
\usepackage{nomencl}
\usepackage[numbers,sort&compress]{natbib}
\usepackage{hyperref}
\usepackage[]{todonotes}

\input{config/math}
\input{config/notation}

\title{Planning with Expected Free Energy in Learned World Models}
\author{%
    Wouter W. L. Nuijten\\
    {\small Eindhoven University of Technology}\\
    {\small Eindhoven, the Netherlands}\\
    {\small Lazy Dynamics B.V., the Netherlands}
    \and
    Bert de Vries\\
    {\small Eindhoven University of Technology}\\
    {\small Eindhoven, the Netherlands}\\
    {\small Lazy Dynamics B.V., the Netherlands}
}
\date{}

\begin{document}
\maketitle
\begin{abstract}
    An agent that learns a model of its environment from scratch has to choose which data to collect.
    When the reward is sparse, the reward is no help with that choice until the agent has found the goal.
    Active inference adds a reason to explore: the expected free energy (EFE) of a plan combines a preference over observations with the information that the plan is expected to give about the agent's model, in one functional.
    So far, EFE-based planners have explored only in models whose structure the agent is given, and most active inference agents with learned neural models train policy networks or drop the information terms.
    We plan with an EFE-based planning functional as it stands, by gradient descent through a learned world model with neural features and a Bayesian last layer, which keeps the information gain in closed form.
    The agent has no value function or learned policy, and the task enters only as a preference prior on observations, whose scale takes the place of an exploration weight; one scale serves all problems.
    We evaluate the same agent, with one configuration and starting from scratch, on four sparse control problems: pendulum swing-up, mountain car, acrobot and cart-pole swing-up.
    In its tenth episode, it keeps the pendulum upright in 75\% of the steps, the tip of the acrobot above the goal height in 21\% and the cart-pole upright in 89\%, more than PETS, PETS with an exploration bonus tuned for each problem, MBPO and SAC.
    Without the epistemic terms, the same agent does not reach the goal in its tenth episode on the pendulum, the mountain car and the acrobot, although a random policy does.
\end{abstract}
\input{sections/introduction}
\input{sections/background}
\input{sections/method}
\input{sections/setup}
\input{sections/experiments-flat}
\input{sections/discussion}
\input{sections/conclusion}
\bibliographystyle{plainnat}
\bibliography{references}
\appendix
\input{appendix/appendix}
\end{document}

%% file: config/notation.tex
\newcommand{\refthm}[1]{\hyperref[#1]{Theorem~\ref*{#1}}}
\newcommand{\reflem}[1]{\hyperref[#1]{Lemma~\ref*{#1}}}
\newcommand{\refappx}[1]{\hyperref[#1]{Appendix~\ref*{#1}}}
\newcommand{\refalgo}[1]{\hyperref[#1]{Algorithm~\ref*{#1}}}
\newcommand{\capsecref}[1]{\hyperref[#1]{Section~\ref*{#1}}}

%% file: sections/introduction.tex
\section{Introduction}\label{sec:introduction}

Sequential decision-making in an unknown environment asks an agent to pursue its goals while it learns how the environment responds to its actions.
Learning and control depend on each other: competent control requires a representation of the environment's state and knowledge of where in that representation the task is solved, yet the agent can learn both only from the experience its own behavior produces.
One of the proposed solutions is reinforcement learning, which learns which actions to take from the reward they accumulate over time \citep{sutton_reinforcement_2018}.
One of the harder problems in reinforcement learning is exploration: the agent learns only about the parts of the environment it visits, so it has to try actions before it knows whether they pay off.
The problem is sharpest when the reward is sparse, since every action earns the same reward until the agent has reached the goal once.
Common remedies inject randomness into action selection or reward the entropy of the policy alongside the task \citep{chua_deep_2018,haarnoja_soft_2018}.
Both are proxies: they make the agent act less predictably in the hope that it reaches parts of the environment it has not seen yet.
Exploration bonuses aim at those parts directly, by adding to the reward an estimate of how much an action would teach the agent about its model \citep{houthooft_vime_2016,pathak_selfsupervised_2019,sekar_planning_2020}.
The bonus needs a weight against the task reward, however, and the best weight changes from task to task.

Active inference, on the other hand, posits that agents plan by minimizing their Expected Free Energy (EFE), a quantity that scores a policy by the beliefs the agent expects to hold after carrying it out \citep{friston_active_2015,parr_active_2022}.
Active inference is motivated by neuroscience, as a theory of how the brain perceives and acts \citep{friston_freeenergy_2010}.
The EFE contains a preference over observations, which plays the role of the reward, and the expected information gain about the agent's model, which plays the role of the exploration bonus.
Both are terms of one functional, and the agent's own uncertainty decides how much exploration is worth.
The weight of the bonus is still there, however: it becomes the scale of the preference, which is part of the agent's model.
Because the information gain depends on the agent's beliefs, the EFE cannot be written as a reward, and for a long time planning with it required search over action sequences \citep{friston_sophisticated_2021}.
Recent work has replaced this exponential tree search with a tractable bound on the EFE, showing promising results in control under uncertainty \citep{nuijten_expected_2026,nuijten_what_2026}.

This line of work has so far only been shown to explore within a known world model.
In these experiments, the agent is given the structure of its model and reduces its uncertainty about which state it is in or about a small set of latent parameters \citep{nuijten_expected_2026,nuijten_what_2026,nuijten_sophisticated_2026}; it never has to learn the world model itself.
Learning the whole model is harder, because the agent has to learn a representation of the state jointly with the dynamics in that representation, and it has to find out where in that representation the preferred observations occur.
Whether minimizing the EFE can guide that learning in continuous control, with a generic function approximator as the model, has not been tested.
Most active inference agents with learned neural models train policy networks, drop the information terms or search over discrete actions \citep{ueltzhoffer_deep_2018,millidge_deep_2020,mazzaglia_contrastive_2021,tschantz_scaling_2020,fountas_deep_2020}; the closest to ours plans in continuous actions with a related functional, the free energy of the expected future \citep{tschantz_reinforcement_2020} (\capsecref{sec:bg-aif}).

In this work, we take the cost functional for plans derived by \citet{nuijten_what_2026}, whose minimizer is the EFE-based plan, to continuous control, and show that it works as a planning objective with a generic learned model.
The agent minimizes this functional as it stands, by gradient descent, with no value function and no learned policy.
Its model of transitions and observations uses neural-network features with a Bayesian last layer, which keeps the information gain in closed form.
We compare the agent with PETS, PETS with the exploration bonus at its best weight for each task, MBPO and SAC, all of which receive the same reward, and with the same agent without the epistemic terms, which maximizes the expected return with the same model and planner.
Our contributions are as follows:
\begin{itemize}
    \item We show that the agent, with one configuration and one scale of the preference for all problems, keeps the pendulum upright in 75\% of the steps of its tenth episode, the acrobot's tip above the goal height in 21\% and the cart-pole upright in 89\%, more than any baseline.
    \item We show that the agent needs the epistemic terms: without them, the agent does not reach the goal in its tenth episode on the pendulum, the mountain car and the acrobot, although a random policy does.
    \item We make explicit that the scale of the preference prior plays the role of an exploration weight, and show that the agent succeeds with every seed from a tenth of this scale to the scale we use, on the three problems we tested, while the best weight of the bonus for PETS changes from problem to problem.
    \item We find that the information gain is high in states the agent has not visited and falls as the model learns, after which the preference takes over, and that performance does not degrade over 100 episodes.
\end{itemize}

\capsecref{sec:background} states the problem and reviews model-based reinforcement learning and active inference\footnote{An interactive explainer with recordings of the agent is at \url{https://wouterwln.github.io/continuousaif/}.}, \capsecref{sec:method} derives the agent, and \capsecref{sec:setup} describes the environments, baselines and protocol.
\capsecref{sec:results} reports the results, and \capsecref{sec:discussion} discusses the limits of the agent and directions for future work.

%% file: sections/background.tex
\section{Background}\label{sec:background}

\subsection{Problem statement}\label{sec:problem}

We consider an agent acting in a partially observed environment in discrete time, in episodes of $H$ steps \citep{kaelbling_planning_1998}.
At step $t$ the agent takes an action $u_t \in \mathcal{U}$, the hidden state moves to $x_t$ according to $p(x_t \mid x_{t-1}, u_t)$, and the agent receives an observation $y_t \in \mathcal{Y}$ drawn from $p(y_t \mid x_t)$.
The reward is part of the observation, $R_t = r(y_t)$ for a known function $r$, so it needs no factor of its own.
Boldface denotes a sequence over the episode, $\bm{x} = (x_0, \dots, x_H)$, $\bm{y} = (y_1, \dots, y_H)$ and $\bm{u} = (u_1, \dots, u_H)$, so that the environment is described by
\begin{equation}\label{eq:pomdp}
    p(\bm{y}, \bm{x} \mid \bm{u}) = p(x_0) \prod_{t=1}^{H} p(y_t \mid x_t)\, p(x_t \mid x_{t-1}, u_t).
\end{equation}
A \emph{plan} $\bm{u}$ is a fixed sequence of actions.
A \emph{policy} $\pi$ chooses each action $u_t$ from the actions and observations before step $t$.
The objective is a policy that maximizes the expected return of an episode,
\begin{equation}\label{eq:return}
    \max_\pi \Ex{\pi}{\sum_{t=1}^{H} R_t}.
\end{equation}

In earlier work on EFE-based planning, the agent knew this model up to a small set of parameters \citep{nuijten_expected_2026,nuijten_what_2026}.
Here it knows neither the transition nor the observation model, and it never observes $x_t$, so it has to learn a representation of the state together with the dynamics and the observation model in that representation.

\subsection{Model-based reinforcement learning}\label{sec:bg-mbrl}

Reinforcement learning approaches this problem in two ways.
Model-free methods learn a policy or a value function directly from the agent's samples \citep{mnih_humanlevel_2015,haarnoja_soft_2018}.
They reach strong final performance in continuous control, but they need many interactions to get there, because every sample is used only to update the policy or the value function.
Model-based methods first learn a model of the environment from the samples and then use it to plan or to generate more training data, so that each interaction teaches the agent more \citep{moerland_modelbased_2023}.

The model-based methods differ mainly in the model they learn and in how they turn it into actions.
PILCO learns a Gaussian process model of the dynamics and improves a policy by propagating its uncertainty analytically, which makes it very data efficient on low-dimensional problems such as cart-pole swing-up \citep{deisenroth_pilco_2011}.
PETS replaces the Gaussian process with an ensemble of probabilistic neural networks and plans directly with the model: at every step, it samples action sequences, scores them by their predicted reward under the ensemble, executes the first action of the best one and replans \citep{chua_deep_2018}.
MBPO uses a similar ensemble to generate short rollouts from visited states, on which it trains a model-free agent \citep{janner_when_2019}.
When the state is not observed, the model is learned in a latent space.
World Models and PlaNet learn latent dynamics from images, and PlaNet plans in that latent space in the same way as PETS \citep{ha_world_2018,hafner_learning_2019}; Dreamer learns a policy and a value function on trajectories imagined by such a model \citep{hafner_dream_2020}.
In these methods, the uncertainty of the model mainly guards the planner against exploiting the errors of the model.
The agent does not seek out the states where the model is uncertain.

Exploration is handled separately.
Most model-based agents explore through the randomness of their planner or the entropy of their policy, as their model-free counterparts do.
Methods aimed at exploration add an intrinsic reward for states that are rarely visited \citep{bellemare_unifying_2016}, for states where a learned predictor is wrong \citep{pathak_curiositydriven_2017,burda_exploration_2018}, or for the information an action is expected to give about the model \citep{houthooft_vime_2016,sekar_planning_2020}, each with a weight against the task reward.
Posterior sampling explores without a bonus, by acting on a sample of the model or the value function that changes from episode to episode \citep{osband_more_2013,osband_deep_2019}.

Our agent is model based in this sense: it learns a latent model of the form \eqref{eq:pomdp}, with parameters $\theta$ in its transition and observation maps, and plans with it at every step, as PETS and PlaNet do (\capsecref{sec:method}).
It represents the uncertainty of the model with a Bayesian last layer \citep{snoek_scalable_2015,riquelme_deep_2018} in place of an ensemble, and its exploration comes from the planning objective, without an intrinsic reward or a weight.

\subsection{Active inference}\label{sec:bg-aif}

Active inference treats perception and action as inference in one generative model \citep{friston_active_2015,parr_active_2022}.
The agent scores a plan by its expected free energy, which splits into the expected negative log-preference of the observations the plan leads to and the information that these observations are expected to give about the hidden states and the parameters of the model \citep{dacosta_active_2020}.
The information gain about the parameters drives the agent to the parts of the environment where its model is uncertain \citep{schwartenbeck_exploration_2013,schwartenbeck_computational_2019}.
Because the information gain depends on the agent's beliefs, the EFE of a plan is not an expected sum of rewards, and dynamic programming over states does not apply to it.

Most EFE planners therefore search.
Early agents enumerated a set of plans and scored each by its EFE \citep{friston_active_2015,dacosta_active_2020}; sophisticated inference searches a tree over future actions and beliefs \citep{friston_sophisticated_2021}, and later work makes this search cheaper \citep{champion_branching_2022,paul_efficient_2024}.
A second line connects active inference to planning as inference, in which the agent infers its actions in a model conditioned on its goals \citep{attias_planning_2003,toussaint_probabilistic_2006,botvinick_planning_2012,levine_reinforcement_2018,millidge_relationship_2020}.
Which plan this inference yields depends on the inference problem: posterior inference with free beliefs over the dynamics is optimistic, and other types of inference give other planners \citep{levine_reinforcement_2018,lazaro-gredilla_what_2024}.
\citet{nuijten_expected_2026} recast EFE minimization as variational free energy minimization in a model with epistemic priors, \citet{nuijten_message_2026} gave a message passing realization, and \citet{nuijten_what_2026} derived the functional $F_{\text{EFE}}$, whose minimizer is the EFE-based plan, by adding a policy-entropy correction and an EFE correction to the variational free energy.
Epistemic priors also yield sophisticated policies \citep{nuijten_sophisticated_2026}.
All of these planners work with models whose structure the agent is given.

Active inference agents with learned neural models make other choices.
Some do not plan with the EFE and instead train a policy network on the free energy of past observations \citep{ueltzhoffer_deep_2018}.
Others amortize the EFE in a policy network and a network that predicts the EFE, trained as in actor-critic methods, with the reward as the preference \citep{millidge_deep_2020} or with a goal image as the preference and no information gain term \citep{mazzaglia_contrastive_2021}.
Those that plan with the EFE use the reward as the preference, and either drop the information terms when the reward is present \citep{tschantz_scaling_2020} or select discrete actions by tree search guided by a learned habit network \citep{fountas_deep_2020}.
\citet{tschantz_reinforcement_2020} plan in continuous actions with the cross-entropy method over an ensemble model and minimize a related functional, the free energy of the expected future, which keeps the information gain about the transition parameters next to a preference over rewards.
Our agent plans with $F_{\text{EFE}}$ as \citet{nuijten_what_2026} derived it, keeps all its terms, and replaces message passing in a given model by gradient descent through a learned one, with continuous states and actions.

%% file: sections/method.tex
\section{Method}\label{sec:method}

\subsection{The generative model}\label{sec:method-model}

The agent models its environment with a state-space model: a latent state $x_t$ evolves under the actions $u_t$ and generates the observations $y_t$, and parameters $\theta$ govern both.
Over a horizon of $T$ steps, the predictive generative model is
\begin{equation}\label{eq:generative-model}
    p(\bm{y}, \bm{x}, \bm{u}, \theta) = p(x_0)\,p(\theta) \prod_{t=1}^{T} p(y_t \mid x_t, \theta)\, p(x_t \mid x_{t-1}, u_t, \theta)\, p(u_t),
\end{equation}
where $p(x_0)$ and $p(\theta)$ are the agent's current beliefs.
In this section, boldface denotes a sequence over the planning horizon, $\bm{x} = (x_0, \dots, x_T)$, instead of the episode.
The environment reaches the agent only through $y_t$, and nothing fixes what the latent state means.
The model is not specific to a task, so the agent has to learn a representation of its environment together with the dynamics and the observation model, from its own data.

We parameterize the transition and observation models with neural networks.
They approximate nonlinear functions \citep{hornik_multilayer_1989} and learn representations from data \citep{bengio_representation_2013}, so the agent can find a latent space and dynamics that suit each environment without hand-designed features, as in learned latent dynamics models for control \citep{hafner_learning_2019}.
The epistemic terms of the planning objective are functionals of the agent's beliefs about the parameters, so the agent explores well only if these beliefs give a usable estimate of its uncertainty.
Neural networks trained to point estimates have no such beliefs, and their predictions are often overconfident on inputs outside the training data distribution \citep{guo_calibration_2017,lakshminarayanan_simple_2017,ovadia_can_2019}.
We obtain an approximation by making only the last layer of each network Bayesian.
Feature networks compute $\phi(x_{t-1}, u_t)$ and $\psi(x_t)$, and the transition and observation models are Gaussian with means linear in these features,
\begin{align}
    p(x_t \mid x_{t-1}, u_t, \theta) & = \mathcal{N}\big(x_t;\; x_{t-1} + W_x^\top \phi(x_{t-1}, u_t),\; \Sigma_x\big),     \\
    p(y_t \mid x_t, \theta)          & = \mathcal{N}\big(y_t;\; W_y^\top \psi(x_t),\; \operatorname{diag}(\sigma_y^2)\big).
\end{align}
Only the last-layer weights $\theta = (W_x, W_y)$ are uncertain; the feature networks and the noise covariances are point estimates.
With a matrix-normal belief $q(\theta)$, with mean $M_x$ and row covariance $S_x$ for the transition weights and $M_y$, $S_y$ for the observation weights, the weights integrate out in closed form,
\begin{align}
    q(x_t \mid x_{t-1}, u_t) & = \int p(x_t \mid x_{t-1}, u_t, \theta)\, q(\theta) \diff\theta = \mathcal{N}\big(x_t;\; x_{t-1} + M_x^\top \phi,\; (1 + \phi^\top S_x \phi)\,\Sigma_x\big), \\
    q(y_t \mid x_t)          & = \int p(y_t \mid x_t, \theta)\, q(\theta) \diff\theta = \mathcal{N}\big(y_t;\; M_y^\top \psi,\; \operatorname{diag}(\sigma_y^2 + \psi^\top S_y \psi)\big),
\end{align}
with $\phi = \phi(x_{t-1}, u_t)$ and $\psi = \psi(x_t)$.
This is Bayesian linear regression on learned features \citep{snoek_scalable_2015,riquelme_deep_2018,harrison_metalearning_2020}.
Under a prior with precision $\alpha$ on the weights, the row covariances are
\begin{equation}\label{eq:row-covariance}
    S_x = \big(\alpha I + \Phi^\top \Phi\big)^{-1}, \qquad
    S_y = \big(\alpha I + \bar{\sigma}_y^{-2}\, \Psi^\top \Psi\big)^{-1},
\end{equation}
where the rows of $\Phi$ and $\Psi$ are the features of the transitions and states in the data, and $\bar{\sigma}_y^{-2}$ is the mean of the $\sigma_{y,j}^{-2}$.
The matrices $S_x$ and $S_y$ summarize the data and do not depend on the state.
The quadratic form $\phi(x_{t-1}, u_t)^\top S_x \phi(x_{t-1}, u_t)$ does: it is small when $\phi(x_{t-1}, u_t)$ points in directions of feature space that the data cover, and close to its prior value $\|\phi(x_{t-1}, u_t)\|^2 / \alpha$ in directions they do not, and the same holds for $\psi(x_t)^\top S_y \psi(x_t)$.
It captures only the uncertainty of the last layer for fixed features, but it costs one quadratic form per prediction, and since the noise covariances are the same at every state, all variation of the predictive variance over the state space comes from $q(\theta)$.
\autoref{fig:uncertainty-maps} in \capsecref{sec:results} shows the resulting information gain for the pendulum during learning.

The task enters the model only through a \emph{preference prior} $\hat{p}(\bm{y}) = \prod_{t=1}^{T} \hat{p}(y_t)$ on observations.
Since the reward is part of the observation, $R_t = r(y_t)$ (\capsecref{sec:problem}), the preference is for high reward, $\hat{p}(y_t) \propto \exp(r(y_t))$.
We give the agent this function, so it learns the dynamics and the observation model but not the reward.
The agent does not know where in its latent space these observations occur or which actions lead there, and has to find out by acting.
We never place preferences on the latent state, whose coordinates are learned and have no fixed meaning.
The agent fits $q(\theta)$, the feature networks and an amortized filter for $x_t$ to all data collected so far, as described in \capsecref{sec:agent}.

\subsection{Planning as inference}\label{sec:method-planning}

The agent plans by minimizing the variational free energy of the preference-augmented model $\hat{p}(\bm{y}, \bm{x}, \bm{u}, \theta) \propto p(\bm{y}, \bm{x}, \bm{u}, \theta)\, \hat{p}(\bm{y})$ with the policy-entropy and EFE corrections \citep{nuijten_what_2026},
\begin{equation}\label{eq:f-efe}
    F_{\text{EFE}}[q] = \FE{\hat{p}}{q}
    + \sum_{t=1}^T \entS{q}{u_t \mid x_{t-1}}
    + \sum_{t=1}^T \Big( 2\,\entS{q}{y_t \mid x_t, \theta} - \entS{q}{x_t \mid x_{t-1}, u_t} - \entS{q}{y_t \mid x_t} \Big),
\end{equation}
over the family
\begin{equation}\label{eq:family}
    q(\bm{y}, \bm{x}, \bm{u}, \theta) = q(x_0)\, q(\theta) \prod_{t=1}^{T} p(y_t \mid x_t, \theta)\, p(x_t \mid x_{t-1}, u_t, \theta)\, q(u_t \mid x_{t-1}).
\end{equation}
The beliefs $q(x_0)$ and $q(\theta)$ come from the filter and stay fixed during planning, the transition and observation factors are those of the model, and only the action posterior $q(u_t \mid x_{t-1})$ is free.
Unlike in \citet{nuijten_what_2026}, the future state beliefs are therefore not free; they follow from the actions through the model.

The transition factor is held at the model because a free one makes the agent optimistic: inference moves probability toward the preferred states and favors plans in which chance carries the agent to its goal \citep{levine_reinforcement_2018,lazaro-gredilla_what_2024}.
The transition noise is a property of the environment that the agent can neither control nor infer by planning, so its transition belief has to be the transition model.

Under \eqref{eq:family}, the policy-entropy correction cancels the $\log q(u_t \mid x_{t-1})$ term in $\FE{\hat{p}}{q}$, the terms in $q(x_0)$ and $q(\theta)$ are constant, and an action prior that is uniform on the action box adds another constant, so
\begin{equation}\label{eq:planning-cost}
    \begin{aligned}
        F_{\text{EFE}}[q] & = \sum_{t=1}^{T} \Ex{q}{\ell_t} + \text{const},                                                                   \\
        \ell_t            & = -\log \hat{p}(y_t) - \ent{q(x_t \mid x_{t-1}, u_t)} + 2\,\ent{p(y_t \mid x_t, \theta)} - \ent{q(y_t \mid x_t)}.
    \end{aligned}
\end{equation}
With the predictive distributions of \capsecref{sec:method-model}, the entropies are closed form,
\begin{align}
    \ent{q(x_t \mid x_{t-1}, u_t)}                        & = \tfrac{1}{2} \log\det(2\pi e \Sigma_x) + \tfrac{d_x}{2} \log\big(1 + \phi(x_{t-1}, u_t)^\top S_x \phi(x_{t-1}, u_t)\big), \\
    \ent{p(y_t \mid x_t, \theta)} - \ent{q(y_t \mid x_t)} & = -\tfrac{1}{2} \textstyle\sum_{j} \log\big(1 + \psi(x_t)^\top S_y \psi(x_t) / \sigma_{y,j}^2\big),
\end{align}
and $\ent{p(y_t \mid x_t, \theta)}$ is constant.
Up to constants, $\ell_t$ is the negative log-preference minus the information gains about the transition weights at $(x_{t-1}, u_t)$ and about the observation weights at $x_t$.

The two information gains come from different terms of $F_{\text{EFE}}$.
The gain about the observation weights is the familiar novelty term of the EFE: $\entS{q}{y_t \mid x_t, \theta} - \entS{q}{y_t \mid x_t} = -\MI{q}{y_t ; \theta \mid x_t}$ is the information that the observation is expected to give about the parameters.
The gain about the transition weights has no such counterpart, since the usual EFE scores only the information that observations carry.
It comes from $-\entS{q}{x_t \mid x_{t-1}, u_t}$, the term that the epistemic prior on the actions contributes to the EFE correction \citep{nuijten_expected_2026,nuijten_what_2026}.
Its entropy is that of the transition marginalized over $\theta$, which splits into the entropy of the transition noise and the information that the next state carries about the parameters,
\begin{equation}\label{eq:transition-split}
    \entS{q}{x_t \mid x_{t-1}, u_t} = \entS{q}{x_t \mid x_{t-1}, u_t, \theta} + \MI{q}{x_t ; \theta \mid x_{t-1}, u_t}.
\end{equation}
The noise entropy is the same for every state and action, because $\Sigma_x$ is, so only the information gain depends on the plan.
The agent never observes $x_t$, so this is information about its latent dynamics, which it obtains through the filter.
Mutual information does not change under invertible maps of $x_t$, so the term does not depend on the coordinates that the agent happens to learn for its latent space.

The functional has no weight between the preference and the information gains, but the scale of the preference acts as one.
A preference with temperature $\tau$, $\hat{p}_\tau(y_t) \propto \exp(r(y_t) / \tau)$, gives the cost $-r(y_t)/\tau$ in place of $-r(y_t)$ in $\ell_t$, and $\tau$ times this cost is the cost of $\hat{p}$ with the information gains multiplied by $\tau$.
Both costs have the same minimizer, so choosing $\tau$ is the same as choosing the weight of an exploration bonus.
In active inference, this weight is part of the generative model: the preference prior fixes, in nats, how much a unit of preference is worth against a unit of information.
The information gains have a scale of their own, set by the prior precision $\alpha$ and the size of the features.
We use $\tau = 1$, with the reward of \refappx{app:environments}, whose preference separates the goal from the rest of the observation space by about 7 nats, and the same $\alpha$ for all four problems; \refappx{app:preference-scale} varies $\tau$ over two orders of magnitude.

The expectation in \eqref{eq:planning-cost} has no closed form, because the nonlinear features make $q(x_t)$ non-Gaussian after the first step and all states of a trajectory share the same $\theta$.
We estimate it with $N$ simulated trajectories.
Each draws $x_0 \sim q(x_0)$ and one $\theta \sim q(\theta)$ for the whole horizon, and then samples $u_t \sim q(u_t \mid x_{t-1})$, $x_t \sim p(x_t \mid x_{t-1}, u_t, \theta)$ and $y_t \sim p(y_t \mid x_t, \theta)$ for $t = 1, \dots, T$.
The cost $\ell_t$ is exact at the sampled points, so sampling only replaces the outer expectation,
\begin{equation}\label{eq:f-efe-mc}
    F_{\text{EFE}}[q] \approx \frac{1}{N} \sum_{n=1}^{N} \sum_{t=1}^{T} \ell_t^{(n)} + \text{const}.
\end{equation}

We approximate the action posterior $q(u_t \mid x_{t-1})$ with a parametric form, so $F_{\text{EFE}}$ becomes a function of the parameters $\lambda$ of that form, and planning becomes a minimization over $\lambda$.
The form is a Gaussian on an unbounded variable $z_t$ whose mean depends linearly on the previous state, squashed into the action box $[-u_{\max}, u_{\max}]$,
\begin{equation}\label{eq:action-posterior}
    q(z_t \mid x_{t-1}) = \mathcal{N}\big(z_t;\; \mu_t + K_t (x_{t-1} - \bar{x}_{t-1}),\; \operatorname{diag}(\sigma_t^2)\big), \qquad
    u_t = u_{\max} \tanh(z_t),
\end{equation}
with $\lambda = \{\mu_t, K_t, \sigma_t\}_{t=1}^{T}$ and $\bar{x}_{t-1}$ the mean of the simulated states at step $t-1$.
The gain $K_t$ lets the plan react to the state of each simulated trajectory, which spreads under the sampled $\theta$ and the transition noise.
Every sample is a differentiable function of $\lambda$ and of standard normal noise that does not depend on $\lambda$ \citep{kingma_autoencoding_2022}, so backpropagation through the simulated trajectories and the world model gives an unbiased estimate of $\nabla_\lambda F_{\text{EFE}}$.
At every environment step the agent takes a few Adam steps \citep{kingma_adam_2015} on $\lambda$ from two initial plans, the previous plan shifted by one step and a fresh plan, and keeps the one with the lower estimate of $F_{\text{EFE}}$ on a common set of simulated trajectories.
The fresh plan lets the agent leave a plan that has settled on the boundary of the action box.
The agent then executes the mode of $q(u_1 \mid x_0)$ at the mean of $q(x_0)$, where the feedback term vanishes, and replans after the next observation, as in receding-horizon control \citep{richalet_model_1978}.
The squashing moves this mode away from $u_{\max}\tanh(\mu_1)$; we find it by Newton's method on $z_1$, and it is unique because the spread is bounded, $\sigma_t < \sigma_{\max}$ with $\sigma_{\max}^2 < 1/2$.
The planning horizon is at most the episode length, $T \leq H$, so the agent plans over a window that moves through the episode.

%% file: sections/setup.tex
\section{Experimental setup}\label{sec:setup}

\subsection{Environments}\label{sec:environments}

\begin{figure}[t]
    \centering
    \includegraphics[width=\textwidth]{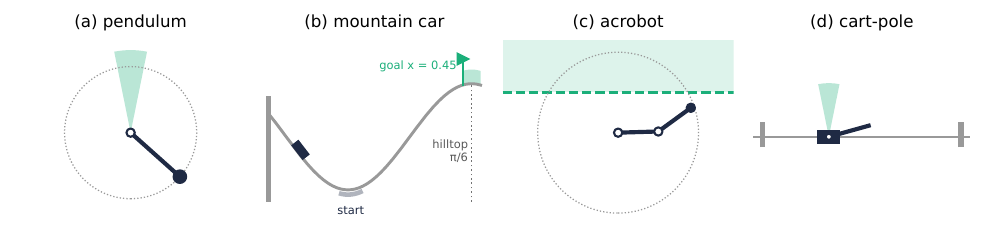}
    \caption{The control problems. Green: goal region.}
    \label{fig:environments}
\end{figure}

We evaluate the agent on four classic control problems: pendulum swing-up, mountain car, acrobot and cart-pole swing-up \citep{brockman_openai_2016} (\autoref{fig:environments}).\footnote{Code for the agent, the baselines and all experiments: \url{https://github.com/biaslab/gradefe}.}
In each of them, the agent starts at rest at the bottom and has to swing to gain energy before it can reach the goal.
The reward is sparse: it is constant everywhere except near the goal, so nothing points the agent to the goal until it has observed it.
\refappx{app:flat} adds dense rewards for the pendulum and the cart-pole.

We measure the fraction of steps that the system spends in the goal region.
Gym ends a mountain-car episode when the car reaches the goal and an acrobot episode when the tip first crosses the line; we run both without termination, so the agent has to stay in the goal region.
\refappx{app:environments} gives the dynamics, observations, rewards and goal regions.

\subsection{Agent}\label{sec:agent}

The agent uses one configuration in all environments: only the episode length differs, and no setting is tuned per task.

The feature networks $\phi$ and $\psi$ of \capsecref{sec:method-model} are multilayer perceptrons with a Bayesian last layer, and the latent space has the same dimension as the observation space.
An amortized filter, a multilayer perceptron that maps a sample of the previous latent state, the action and the new observation to a belief over the current latent state, gives the agent its belief $q(x_0)$ at every step.

The agent starts without data, with random feature networks and the prior over the last-layer weights.
At a fixed interval of environment steps, it trains the feature networks and the filter on all data collected so far, by gradient ascent on the evidence lower bound of the observed sequences.
It then sets $q(\theta)$ and the noise covariances in closed form for the trained features, since the model is conjugate in the last layer; for the observation weights, the update uses the mean noise precision of \eqref{eq:row-covariance} and is therefore approximate.
Retraining changes the latent space, so when an update falls inside an episode, the agent reruns the new filter over the observations of the episode so far to obtain its current belief $q(x_0)$.
Between these updates, the model does not change.

At every environment step, the agent takes a fixed number of gradient steps on $F_{\text{EFE}}$ (\capsecref{sec:method-planning}) and acts.
The number is small, so the plan of one step is not converged; because each plan starts from the previous one, the gradient steps accumulate over the episode.
\refappx{app:agent} lists all settings.

\subsection{Baselines}\label{sec:baselines}

To isolate the epistemic terms, we also run the agent without them.
This preference-only agent minimizes $F_{\text{EFE}}$ without its EFE correction, the last sum in \eqref{eq:f-efe}, which leaves the variational free energy with the policy-entropy correction of planning as inference \citep{lazaro-gredilla_what_2024}.
Under the family \eqref{eq:family}, its cost is the expected negative log preference $\sum_{t=1}^{T} \Ex{q}{-\log \hat{p}(y_t)}$, so it maximizes the expected return with the same model and planner as our agent.
The EFE correction also contains the ambiguity $\ent{p(y_t \mid x_t, \theta)}$, which is the same in every state under our model, so the two agents differ only in the information gains.

We compare the agent with model-based and model-free reinforcement learning.
PETS \citep{chua_deep_2018} plans with an ensemble of probabilistic networks and the cross-entropy method; it is the closest model-based planner without an epistemic term.
We give its networks the size of our feature networks (three hidden layers of 64) instead of the published three layers of 500, so that both agents have the same capacity per network; PETS has an ensemble of five.
MBPO keeps its published networks.
To test whether an exploration bonus closes the gap, we also run PETS with a bonus for the disagreement of its ensemble \citep{pathak_selfsupervised_2019,sekar_planning_2020}, with the weight of the bonus chosen per task.
MBPO \citep{janner_when_2019} learns a policy from short rollouts of a learned model, and SAC \citep{haarnoja_soft_2018} learns a policy from the environment data only.

The baselines solve the same problems as our agent, with the same episode lengths, initial states and success test.
Our agent and PETS act without added noise, with the mode of $q(u_1 \mid x_0)$ and the first action of the final CEM mean, so we score them on their training episodes; PETS with the bonus keeps the bonus in these episodes, as our agent keeps its epistemic terms.
SAC and MBPO sample their actions during training, so after every training episode we run one evaluation episode with the mean action $\tanh(\mu)$, without learning, from the same initial state, and score them on it.
They receive the same reward $R_t = r(y_t)$, without shaping, and PETS and MBPO use the function $r$ itself, as our agent does.
We reimplemented the baselines in JAX, following the code of their authors.
\refappx{app:baselines} lists their settings and where they depart from the published ones.

\subsection{Protocol}\label{sec:protocol}

Every run starts from scratch, without data and with random networks, and its seed fixes the initial networks and the noise of the environment.
Every method runs 10 episodes on each of 10 seeds.
Ten episodes are 3000 to 5000 environment steps, a budget in which model-free methods are not expected to solve these problems; we include SAC as a reference point for model-free learning in this regime.
We summarize each comparison of two methods by the probability of superiority \citep{agarwal_deep_2021}: the probability that a randomly chosen seed of one method spends more steps of its tenth episode in the goal region than a randomly chosen seed of the other, with ties counted as one half.
\refappx{app:flat} gives two-sided Mann--Whitney U tests on the same values, with a Holm correction where one claim rests on several tests.
We fixed the conditions, seeds, metrics and tests of the main comparisons before we ran them.
The one exception is the metric, which we changed after the runs (\capsecref{sec:results}).

%% file: sections/experiments-flat.tex
\section{Results}\label{sec:results}

In the tenth episode, our agent keeps the pendulum upright in 75\% of the steps, the acrobot above the line in 21\% and the cart-pole upright in 89\% (\autoref{tab:flat}).
The strongest baseline is PETS with the exploration bonus at the weight that is best for each problem (10, 1, 0.1 and 0 for the pendulum, the mountain car, the acrobot and the cart-pole), which reaches 27\% on the pendulum and 13\% on the acrobot; our agent has no such weight to choose.
Paired seed by seed with this baseline, our agent spends more steps in the goal region in 90\% of the pairs on the pendulum and in 88\% on the acrobot, and against every other baseline on these three problems it wins at least 91\% of the pairs.
On the acrobot, our agent swings the tip above the line again and again but does not balance it there.
On the mountain car, the fraction of steps does not separate the methods.
Without Gym's termination, the goal region contains an absorbing state, and a random policy spends 32\% of its steps there (\refappx{app:mountain-car}); our agent spends 38\% and PETS with the bonus 33\%, and our agent wins 61\% of the pairs.
Before the runs, we had fixed success counts in the tenth episode as the metric instead of this fraction, but on the acrobot a random policy meets that success criterion in most episodes; \refappx{app:flat} gives the counts, which lead to the same conclusions, except that the difference to PETS with the bonus is not significant on the acrobot and is significant on the mountain car.

\begin{table}[t]
    \centering
    \caption{Fraction of steps in the goal region in the tenth episode, mean $\pm$ standard deviation over 10 seeds. An episode has $H$ steps.}
    \label{tab:flat}
    \footnotesize
    \setlength{\tabcolsep}{3pt}
    \begin{tabular}{@{}lcccccccc@{}}
        \toprule
        Problem      & $H$ & Ours                     & Pref.\ only              & PETS            & PETS + bonus        & MBPO            & SAC             & Random \\
        \midrule
        Pendulum     & 300 & \textbf{0.75} $\pm$ 0.13 & 0.00 $\pm$ 0.00          & 0.00 $\pm$ 0.00 & 0.27 $\pm$ 0.34     & 0.23 $\pm$ 0.30 & 0.00 $\pm$ 0.00 & 0.03   \\
        Mountain car & 400 & 0.38 $\pm$ 0.24 & 0.00 $\pm$ 0.00          & 0.00 $\pm$ 0.00 & 0.33 $\pm$ 0.34     & 0.21 $\pm$ 0.35 & 0.00 $\pm$ 0.00 & 0.32   \\
        Acrobot      & 500 & \textbf{0.21} $\pm$ 0.04 & 0.00 $\pm$ 0.00          & 0.03 $\pm$ 0.04 & 0.13 $\pm$ 0.07     & 0.02 $\pm$ 0.05 & 0.00 $\pm$ 0.00 & 0.03   \\
        Cart-pole    & 400 & \textbf{0.89} $\pm$ 0.04 & \textbf{0.89} $\pm$ 0.03 & 0.42 $\pm$ 0.18 & 0.42 $\pm$ 0.18$^*$ & 0.30 $\pm$ 0.36 & 0.08 $\pm$ 0.11 & 0.05   \\
        \bottomrule
    \end{tabular}

    \smallskip
    {\footnotesize $^*$The best bonus weight on the cart-pole is 0, which is PETS. MBPO and SAC: greedy evaluation episode after training episode 10. Random: bang-bang actions held for 10 steps, 1000 episodes.}
\end{table}

\begin{figure}[t]
    \centering
    \includegraphics[width=\textwidth]{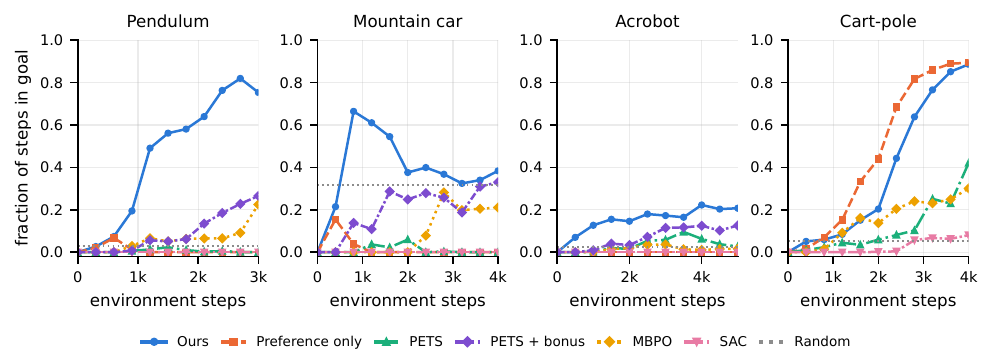}
    \caption{Fraction of steps in the goal region per episode, mean over 10 seeds, against environment steps. Each point after the origin is the end of an episode; for MBPO and SAC, the greedy evaluation episode after it. Random as in \autoref{tab:flat}.}
    \label{fig:curves}
\end{figure}

\autoref{fig:curves} shows how many interactions the agents need.
On the pendulum, the acrobot and the cart-pole, our agent is above the random policy from the second episode on, and it keeps the pendulum upright in more than half of the steps after 1500 interactions and the cart-pole after 2800.
Within the same 10 episodes, no baseline exceeds 42\% on any problem.

The epistemic terms make the difference on the pendulum, the mountain car and the acrobot.
Without them, the agent never reaches the goal region in the tenth episode, so the full agent wins every pair, although a random policy reaches the goal region on all three.
When the goal is outside its horizon, the preference is flat over all trajectories that the agent can simulate, so $F_{\text{EFE}}$ has no gradient in the actions and the agent stays near its start.
By the tenth episode, the full agent has visited 58\% and 17\% of the cells of a grid over the state space of the pendulum and the acrobot, and the agent without the epistemic terms 26\% and 0.2\%.
The cart-pole is the exception: a policy with uniformly random actions swings the light pole through the upright position in half of its episodes, so the agent observes the goal without exploring, and both agents reach 89\%.

\begin{figure}[t]
    \centering
    \begin{subfigure}[b]{0.28\textwidth}
        \centering
        \includegraphics[width=\textwidth]{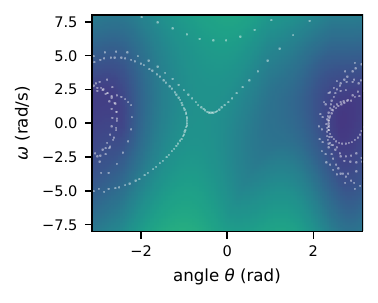}
        \caption{After 1 episode.}
    \end{subfigure}\hfill
    \begin{subfigure}[b]{0.28\textwidth}
        \centering
        \includegraphics[width=\textwidth]{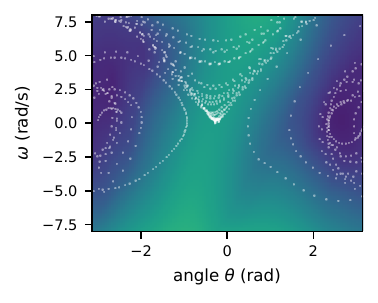}
        \caption{After 3 episodes.}
    \end{subfigure}\hfill
    \begin{subfigure}[b]{0.28\textwidth}
        \centering
        \includegraphics[width=\textwidth]{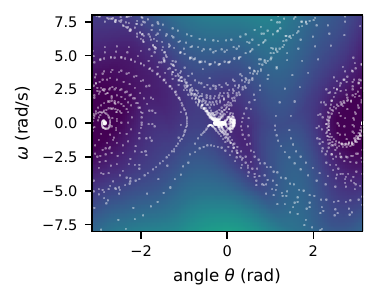}
        \caption{After 10 episodes.}
    \end{subfigure}\hfill
    \begin{subfigure}[b]{0.11\textwidth}
        \centering
        \includegraphics[height=2.9cm]{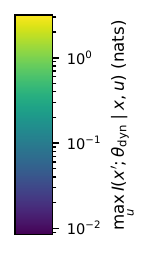}
        \caption*{}
    \end{subfigure}
    \caption{Information gain about the transition weights over the state space of the pendulum, maximized over actions, after 1, 3 and 10 episodes (seed 0). White dots are the visited states.}
    \label{fig:uncertainty-maps}
\end{figure}

\autoref{fig:uncertainty-maps} shows how the information gain about the transition weights changes on the pendulum.
After the first episode, its median is 0.038 nats over the grid cells that the agent has visited and 0.105 nats over the cells it has not.
The agent then goes to the unvisited states with high angular velocity, where it pumps energy into the pendulum.
As the model learns, the information gain falls everywhere, and once the simulated trajectories reach the goal, the preference dominates the plan.
In the tenth episode, the epistemic terms add about 0.03 nats per step, while the negative log preference falls from 6.9 to 0 within 160 steps.

The agent keeps what it has learned.
In runs of 100 episodes, it spends 82\% of the steps of episodes 91 to 100 in the goal region on the pendulum, 90\% on the cart-pole, 28\% on the acrobot and 28\% on the mountain car, against 73\%, 83\%, 21\% and 25\% in the tenth episode of the same runs.

%% file: sections/discussion.tex
\section{Discussion}\label{sec:discussion}

Our results are a proof of concept.
They show that minimizing the EFE gathers the data that a neural world model needs to learn a sparse control task, with no exploration weight, on four problems with low-dimensional observations and one configuration.
They do not show a method that scales, and the clearest obstacle is how the agent learns its model.
Every 100 environment steps, it retrains its feature networks and its filter on all data collected so far, recomputes $q(\theta)$ and reruns its filter over the current episode (\capsecref{sec:agent}).
The interval, the number of gradient steps and the length of the training sequences are set by hand.
Between retrainings, the model does not change, so an observation that contradicts the model affects the plan only after the next retraining, up to 100 steps later.
Because the model is conjugate in its last layer, $q(\theta)$ could follow every observation in closed form between retrainings, with the features held fixed.
We have not tried this, and a learning scheme in which the model and the information gain respond to each observation is the first change we would make.

The planner and the model have further limits.
The agent does not minimize $F_{\text{EFE}}$ to convergence at each step: ten gradient steps per action leave a decrease of 0.1 to 4.3 nats that further steps would find, less than the standard deviation of one 16-trajectory estimate of $F_{\text{EFE}}$ (1.4 to 7.6 nats), and the warm start carries the remaining optimization over to the next steps.
With five gradient steps, the agent solved the pendulum and the mountain car with all three seeds we tried, in about 28\% less time per episode, and the second initial plan wins only 0 to 3\% of the selections after ten steps.
The world model is smooth and has the same noise in every state.
On a cart-pole whose track has walls close to the cart, no method, ours included, succeeded with any of 6 seeds, because the model cannot represent the inelastic collision with the wall, whose steps are 18\% of the data and 62\% of the error in the predicted cart velocity.

Finally, we give the agent the reward function, so it learns the dynamics and the observation model but not its preference prior.
We also choose the scale of that prior, which sets the balance between the preference and the information gains (\capsecref{sec:method-planning}).
One scale served all four problems, and a range of scales served three of them (\refappx{app:preference-scale}), but these problems have low-dimensional observations and the same form of reward.
The information gains grow with the dimensions of the latent and observation spaces, and the preference does not, so we expect the scale that balances them to change on larger problems.

The form of the information gain is a second direction for future work.
The planning cost sums an information gain about the parameters over the steps of a plan, each computed under the same $q(\theta)$ (\capsecref{sec:method-planning}).
This per-step form follows from the Bethe factorization of the variational family in \citet{nuijten_what_2026}.
It counts the information of a region once for every step that the plan spends there, although the second visit would teach the agent much less than the first.
The information that a whole trajectory carries about $\theta$ does not have this problem, and we expect a planning objective without the Bethe assumption, in which the information gain is taken jointly over the trajectory, to explore more efficiently.

A third direction is hierarchy.
The planning horizon limits which goals the agent can see: when the goal is more than $T$ steps away, the preference is flat over all simulated trajectories, and only the information gain shapes the plan.
In active inference, a hierarchy of the same units is a natural extension \citep{devries_active_2026}: an upper unit acts by setting the preference prior of the unit below and observes that unit's observation when it reaches its subgoal or gives up.
Such a hierarchy is short of data: an upper unit makes one decision per option of the unit below, so the number of transitions it learns from falls with every level by the mean option length, and in our preliminary tests a third level did not learn within 100 episodes.
Deeper hierarchies need upper units that learn from fewer decisions, for example from the observations within an option.

%% file: sections/conclusion.tex
\section{Conclusion}\label{sec:conclusion}

We planned with the functional $F_{\text{EFE}}$ of \citet{nuijten_what_2026} as it stands, by gradient descent through a world model of neural features with a Bayesian last layer, with a preference prior on observations whose scale takes the place of an exploration weight, and without a value function or a learned policy.
With one configuration, the agent learned to reach the goal on four sparse control problems from scratch.
In the tenth episode, it kept the pendulum upright in 75\% of the steps, the tip of the acrobot above the goal height in 21\% and the cart-pole upright in 89\%, more than PETS, PETS with an exploration bonus at its best weight for each problem, MBPO and SAC.
Without the epistemic terms, the same agent did not reach the goal in its tenth episode on the pendulum, the mountain car and the acrobot, which a random policy reaches.

%% file: appendix/appendix.tex
\section{Environments}\label{app:environments}

All environments are our own JAX implementations.
Every environment adds small process noise to its velocities and Gaussian noise with standard deviation $0.01$ to every observation, and the agent never sees the true state.
The four control problems follow the equations of their Gym versions \citep{brockman_openai_2016}, and their observations encode angles by their cosine and sine and divide velocities by fixed constants.
\autoref{tab:environments} lists the observations, the action dimensions, the episode lengths and the goal regions.

The sparse reward is
\begin{equation}\label{eq:sparse-reward}
    r(y) = \log\Big(\epsilon + \exp\Big(-\frac{d(y)^2}{2 s^2}\Big)\Big),
\end{equation}
where $d(y)$ is the distance of the observation to the goal set, for example $1 - \cos\theta$ for the upright pendulum, $s$ is the width of the goal region and $\epsilon = 10^{-3}$.
Away from the goal, the reward is constant at $\log \epsilon$, so it gives the agent no direction until the agent has observed the goal once.
With this reward, the preference $\hat{p}(y) \propto \exp(r(y))$ is a Gaussian bump on a flat floor.
For the pendulum and the cart-pole, the dense reward increases linearly toward the goal.

The main text reports the fraction of steps in the goal region.
The success criteria that we fixed before the runs, used in \refappx{app:flat}, are the following: for the pendulum and the cart-pole, the true state has to be inside the goal region in at least 80\% of the last 100 steps of the episode; the mountain car and the acrobot have to reach the goal once, which is when Gym ends their episodes.

\begin{table}[h]
    \centering
    \caption{Environments. $H$ is the episode length.}
    \label{tab:environments}
    \small
    \begin{tabular}{@{}llcccl@{}}
        \toprule
        Environment      & Observation $y$                                        & $d_y$ & $d_u$ & $H$      & Goal region                \\
        \midrule
        Pendulum         & $\cos\theta, \sin\theta, \dot\theta/8$                 & 3     & 1     & 300      & $|\theta| < 0.2$ (held)    \\
        Mountain car     & $x, \dot x/0.07$                                       & 2     & 1     & 400      & $x \geq 0.45$ (reached)    \\
        Cart-pole        & $x, \dot x/2, \cos\theta, \sin\theta, \dot\theta/8$    & 5     & 1     & 400      & $|\theta| < 0.2$ (held)    \\
        Acrobot          & $\cos\theta_{1,2}, \sin\theta_{1,2}, \dot\theta_1/4\pi, \dot\theta_2/9\pi$ & 6     & 1     & 500      & tip height $> 1$ (reached) \\
        \bottomrule
    \end{tabular}
\end{table}

\section{Agent settings}\label{app:agent}

\autoref{tab:agent} lists the settings of the agent.
They are the same in every environment, except the episode length and the size of the replay buffer.

\begin{table}[h]
    \centering
    \caption{Agent settings.}
    \label{tab:agent}
    \small
    \begin{tabular}{@{}lll@{}}
        \toprule
        Part                   & Setting                                    & Value                                                \\
        \midrule
        World model            & latent dimension $d_x$                     & $d_y$                                                \\
                               & feature networks $\phi$, $\psi$            & 3 hidden layers of 64, SiLU, 32 features             \\
                               & prior precision $\alpha$ of the last layer & 1                                                    \\
                               & noise covariances $\Sigma_x$, $\sigma_y$   & learned, initial standard deviation 0.1 \\
                               & filter                                     & 2 hidden layers of 64, SiLU, 32 sample paths          \\
                               & prior $p(x_0)$                             & $\mathcal{N}(0, I)$                                  \\
        \midrule
        Learning               & interval                                   & every 100 environment steps                          \\
                               & gradient steps per update                  & 250 (Adam, learning rate $10^{-3}$, clip 10)         \\
                               & minibatch                                  & 16 sequences of 50 steps                             \\
                               & replay buffer                              & all episodes (10, or 100 in the long runs)           \\
        \midrule
        Planning               & horizon $T$                                & 40                                                   \\
                               & simulated trajectories $N$                 & 16                                                   \\
                               & gradient steps per environment step        & 10 (100 at the first step of an episode)             \\
                               & optimizer                                  & Adam, learning rate 0.05, clip 10                    \\
                               & initial plans                              & 2, selected on 128 common trajectories               \\
                               & action spread                              & $\sigma_{\max} = 0.7$, initial $\sigma_t = 0.5$      \\
                               & executed action                            & mode of $q(u_1 \mid x_0)$, 20 Newton steps           \\
                               & precision                                  & float32                                              \\
        \bottomrule
    \end{tabular}
\end{table}

\section{Baselines}\label{app:baselines}

All baselines use the problems, episode lengths, initial states and success tests of \refappx{app:environments}, and the reward $r(y_t)$.
We reimplemented PETS, MBPO and SAC in JAX, following the code of their authors.
\autoref{tab:baselines} gives their settings.

\begin{table}[h]
    \centering
    \caption{Baseline settings. Departures from the published settings are marked with $^\dagger$.}
    \label{tab:baselines}
    \small
    \begin{tabular}{@{}lp{0.72\textwidth}@{}}
        \toprule
        Method          & Settings                                                                                                                   \\
        \midrule
        PETS            & 5 probabilistic networks, 3 hidden layers of 64$^\dagger$, swish; 20 particles; CEM with 400 samples, 40 elites, 5 iterations, horizon 25; retrained on all data after every episode \\
        PETS + bonus    & PETS with the ensemble disagreement as a bonus; weight from $\{0, 0.1, 1, 10\}$, chosen per task on separate seeds        \\
        MBPO            & 7 networks, 5 elites, 4 hidden layers of 200; rollouts of length 1; 10 SAC updates per step; known reward function $r$ instead of a learned one$^\dagger$ \\
        SAC             & 2 hidden layers of 256, ReLU; learning rate $3 \cdot 10^{-4}$, batch 256, $\gamma = 0.99$; automatic entropy temperature    \\
        \bottomrule
    \end{tabular}
\end{table}

\section{Additional results of the flat agent}\label{app:flat}

Two-sided Mann--Whitney U tests on the fraction of steps in the goal region in the tenth episode agree with the probabilities of superiority in \capsecref{sec:results}.
Our agent is better than every baseline on the pendulum, the acrobot and the cart-pole ($p < 0.008$ after a Holm correction over the 11 comparisons), its difference to PETS with the bonus on the mountain car is not significant ($p = 0.43$), and without the epistemic terms it is worse on the pendulum, the mountain car and the acrobot ($p < 10^{-4}$ for each).

\autoref{tab:prereg-flat} gives the success counts in the tenth episode, with the criteria of \refappx{app:environments}, for the sparse problems of the main text and for the dense rewards.
Our agent is better than the best of PETS, MBPO and SAC on every sparse problem (two-sided Fisher exact test, $p \leq 0.011$).
Against PETS with the bonus at its best weight, it is better on the pendulum and the cart-pole (Holm $p = 0.012$) and, with 10 more seeds, on the mountain car (20/20 against 10/20, Holm $p = 0.0017$), but not on the acrobot (20/20 against 18/20).
With the dense rewards, PETS comes close to our agent, and the differences are not significant (10/10 against 6/10, $p = 0.087$, and 9/10 against 7/10).

\begin{table}[h]
    \centering
    \caption{Successes out of 10 seeds in the tenth episode, under the criteria fixed before the runs (\refappx{app:environments}). PETS + bonus uses the best bonus weight for each problem. MBPO and SAC: greedy evaluation episode on the sparse problems, training episode on the dense ones.}
    \label{tab:prereg-flat}
    \small
    \begin{tabular}{@{}lccccccc@{}}
        \toprule
        Problem              & $H$ & Ours           & Preference only & PETS & PETS + bonus & MBPO & SAC  \\
        \midrule
        Pendulum, sparse     & 300 & \textbf{10/10} & 0/10            & 0/10 & 3/10         & 2/10 & 0/10 \\
        Mountain car, sparse & 400 & \textbf{10/10} & 0/10            & 0/10 & 6/10         & 3/10 & 0/10 \\
        Acrobot, sparse      & 500 & \textbf{10/10} & 0/10            & 4/10 & 9/10         & 2/10 & 0/10 \\
        Cart-pole, sparse    & 400 & \textbf{10/10} & \textbf{10/10}  & 3/10 & 3/10         & 2/10 & 0/10 \\
        \midrule
        Pendulum, dense      & 300 & \textbf{10/10} & \textbf{10/10}  & 6/10 & --           & 4/10 & 0/10 \\
        Cart-pole, dense     & 400 & 9/10           & \textbf{10/10}  & 7/10 & --           & 1/10 & 0/10 \\
        \bottomrule
    \end{tabular}
\end{table}

\subsection{Mountain car}\label{app:mountain-car}

\begin{figure}[h]
    \centering
    \includegraphics[width=0.46\textwidth]{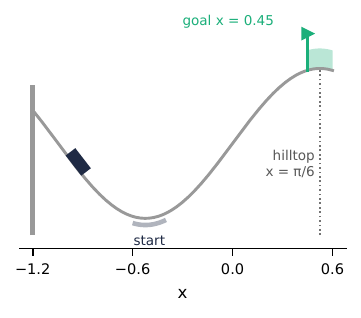}
    \caption{Mountain car. The goal region starts at $x = 0.45$, below the hilltop at $x = \pi/6$.}
    \label{fig:mountain-car}
\end{figure}

Our mountain car has the dynamics of Gym's continuous mountain car, but Gym ends the episode when the car reaches $x \geq 0.45$, and we do not.
Without this termination, the goal region contains an absorbing state (\autoref{fig:mountain-car}).
Gym clips the position at the right bound $x = 0.6$ without setting the velocity to zero, and beyond the hilltop at $x = \pi/6 \approx 0.52$ the slope pulls the car to the right, so a car that passes the hilltop stays at the right bound unless it drives left.
A random policy with bang-bang actions held for 10 steps reaches the goal region in 72\% of its episodes and then spends 32\% of all steps there.
The fraction of steps in the goal region therefore does not measure control on the mountain car, and the reach-once criterion of Gym does not separate exploration from persistent random actions.

On the reach-once criterion, our agent succeeds with every seed (\autoref{tab:prereg-flat}), and it first reaches the goal during the first episode, after a median of 284 steps, when it has refit its model two or three times.
In the tenth episode, it spends 38\% of the steps in the goal region, less than the 66\% of the second episode: its preference peaks at $x = 0.45$, on the slope below the hilltop, and the agent tries to stop there instead of resting at the bound.
Without the epistemic terms, the agent reaches the goal with 5 of 10 seeds in the first episode, while its model is still random, and never after the second episode.
The mountain car therefore supports only the result that the agent without the epistemic terms stops exploring once its model is accurate.

\section{Scale of the preference}\label{app:preference-scale}

\capsecref{sec:method-planning} shows that a preference with temperature $\tau$, $\hat{p}_\tau(y) \propto \exp(r(y)/\tau)$, has the same minimizer of $F_{\text{EFE}}$ as the preference with $\tau = 1$ and the information gains multiplied by $\tau$.
We ran the agent in this second form, with all other settings as in \autoref{tab:agent}, for $\tau \in \{0, 0.1, 0.3, 1, 3, 10\}$ on the sparse pendulum, mountain car and cart-pole, with 6 seeds each.
The value $\tau = 0$, at which the information gains drop out, is the preference-only agent of \capsecref{sec:baselines}, and the runs with $\tau = 0$ and $\tau = 1$ are seeds 0 to 5 of the main results.
We fixed the values of $\tau$, the seeds and the success criterion before the runs, together with the prediction that $\tau = 1$ would be among the best values on every problem.
We did not run the acrobot.

\autoref{tab:preference-scale} gives the results.
From $\tau = 0.1$ to $\tau = 1$, the agent succeeds with every seed on all three problems, and its fraction of steps in the goal region does not change beyond the spread over seeds.
With larger $\tau$, the preference is weaker relative to the information gain, and on the pendulum and the cart-pole some seeds keep exploring instead of holding the goal: with $\tau = 10$, 4 and 2 of 6 seeds succeed.
The mountain car, whose goal region contains an absorbing state (\refappx{app:mountain-car}), is not sensitive to $\tau$.
The agent therefore depends on the scale of its preference as a bonus method depends on its weight, but one scale serves all problems, and every value from 0.1 to 1 works on all of them.
For PETS with the bonus, the weights that work best on the pendulum and the mountain car, 1 and 10, make it fail on the cart-pole with all 6 seeds, where its best weight is 0.
This comparison is indirect: the two methods differ in more than the weight, and we did not run PETS with the bonus over the same range.

\begin{table}[h]
    \centering
    \caption{Scale of the preference. Fraction of steps in the goal region in the tenth episode, mean $\pm$ standard deviation over 6 seeds, and the number of seeds that meet the success criterion of \refappx{app:environments}. The temperature $\tau = 1$ is the configuration of the main results, and $\tau = 0$ is the preference-only agent.}
    \label{tab:preference-scale}
    \small
    \setlength{\tabcolsep}{4pt}
    \begin{tabular}{@{}lcccccc@{}}
        \toprule
        Problem      & $\tau = 0$        & $0.1$             & $0.3$             & $1$               & $3$               & $10$              \\
        \midrule
        Pendulum     & 0.00 $\pm$ 0.00   & 0.79 $\pm$ 0.10   & 0.80 $\pm$ 0.01   & 0.71 $\pm$ 0.16   & 0.66 $\pm$ 0.33   & 0.53 $\pm$ 0.41   \\
                     & 0/6               & 6/6               & 6/6               & 6/6               & 5/6               & 4/6               \\
        Mountain car & 0.00 $\pm$ 0.00   & 0.30 $\pm$ 0.24   & 0.47 $\pm$ 0.24   & 0.33 $\pm$ 0.23   & 0.28 $\pm$ 0.14   & 0.40 $\pm$ 0.21   \\
                     & 0/6               & 6/6               & 6/6               & 6/6               & 6/6               & 6/6               \\
        Cart-pole    & 0.89 $\pm$ 0.04   & 0.85 $\pm$ 0.06   & 0.88 $\pm$ 0.04   & 0.90 $\pm$ 0.02   & 0.77 $\pm$ 0.20   & 0.52 $\pm$ 0.31   \\
                     & 6/6               & 6/6               & 6/6               & 6/6               & 5/6               & 2/6               \\
        \bottomrule
    \end{tabular}
\end{table}